%% file: main-2667-Choi.tex
\documentclass[11pt,a4paper]{article}
\usepackage{times,latexsym}
\usepackage{url}
\usepackage[T1]{fontenc}

\newcommand{\commentout}[1]{}

\usepackage[acceptedWithA]{tacl2018v2}
\usepackage{stmaryrd}
\usepackage{bbm}
\usepackage{enumitem}

\usepackage[]{tacl2018v2}
\usepackage{subcaption}
\usepackage{graphicx}
\usepackage{lipsum}
\usepackage{ifthen}
\usepackage{color}
\usepackage{amsmath}
\usepackage[utf8]{inputenc}
\usepackage{url}
\usepackage{multicol}
\usepackage{tikz}
\usepackage{booktabs}
\usepackage{csquotes}
\usepackage{soul}
\usepackage{float}
\usetikzlibrary{fit}
\usetikzlibrary{patterns}
\usepackage[font=small]{caption}
\usepackage{multirow}

\usepackage{xspace,mfirstuc,tabulary}

\usepackage{xcolor}

\colorlet{centity1}{red!20}
\colorlet{centity2}{orange!20}
\colorlet{centity3}{orange!20}
\colorlet{canswer}{blue!20}
\colorlet{csent}{gray!20}

\newcommand{\sspan}[2][yellow]{%
\sethlcolor{#1}
  {\hl{#2}}
}

\newcommand{\addedcontent}[1]{{#1}}

\newcommand{\spanone}[1]{\sspan[centity1]{#1}}
\newcommand{\spansent}[1]{\sspan[csent]{#1}}

\newcommand\addonly[1]{$\lbag$\colorbox{green!20}{+#1}$\rbag$}

\newcommand\deleteonly[1]{$\lbag$\colorbox{red!20}{-#1}$\rbag$}
\newcommand\swaponeone[2]{$\lbag$\colorbox{red!20}{-#1},\colorbox{green!20}{+#2}$\rbag$}

\newcommand\examplebox[1]{\fbox{\begin{minipage}{.46\textwidth}\footnotesize{#1}\end{minipage}}}

\newcommand{\catt}[1]{$\textsc{#1}$}

\newcommand{\fpr}[3]{#1 {\scriptsize (#2/#3)}}

\newif\iftaclinstructions
\taclinstructionsfalse 
\iftaclinstructions
\renewcommand{\confidential}{}
\renewcommand{\anonsubtext}{(No author info supplied here, for consistency with
TACL-submission anonymization requirements)}
\newcommand{\instr}
\fi

\iftaclpubformat 

\else

\fi

\newtheorem{definition}{Definition}

\title{
Decontextualization: Making Sentences Stand-Alone
}

\author{
Eunsol Choi$^2$\thanks{$\;$Work done at Google.}$\,$, Jennimaria Palomaki$^1$, Matthew Lamm$^1$,\\
{\bf 
Tom Kwiatkowski$^1$, Dipanjan Das$^1$, Michael Collins$^1$}\\
\\
$^1$Google Research \\
$^2$Department of Computer Science, The University of Texas at Austin\\
  {\small {\sf eunsol@cs.utexas.edu, \{jpalomaki, mrlamm, tomkwiat, dipanjand, mjcollins\}@google.com, }} 
}
\date{}

\begin{document}
\maketitle

\begin{abstract}
Models for question answering, dialogue agents, and summarization often interpret the meaning of a sentence in a rich context and use that meaning in a new context.
Taking excerpts of text can be problematic, as key pieces may not be explicit in a local window. We isolate and define the problem of sentence decontextualization: taking a sentence together with its context and rewriting it to be interpretable out of context, while preserving its meaning. We describe an annotation procedure, collect data on the Wikipedia corpus, and use the data to train models to automatically decontextualize sentences. We present preliminary studies that show the value of sentence decontextualization in a user facing task, and as preprocessing for systems that perform document understanding. We argue that decontextualization is an important subtask in many downstream applications, and that the definitions and resources provided can benefit tasks that operate on sentences that occur in a richer context.
\end{abstract}

\input{introduction.tex}
\input{task.tex}
\input{data.tex}
\input{evaluation.tex}

\input{userstudy.tex}

\input{applications.tex}
\input{related.tex}
\input{conclusion.tex}

\section*{Acknowledgments}
We would like to thank members of Google AI, especially Jacob Eisenstein, Kenton Lee, Santiago Ontanon, Ankur Parikh, Daniel Andor, Chris Alberti, Slav Petrov for helpful discussions and comments. Lastly, we would like to thank our annotators.
\bibliography{references.bib}
\bibliographystyle{acl_natbib}

\end{document}

%% file: introduction.tex
\section{Introduction}\label{sec:intro}
\commentout{Language encodes information, and we transfer such information from one context to another. We study how to transfer information in a sentence from a rich context (long document) to a simple context when the sentence is presented alone. We name this task as decontextualization, i.e., editing a sentence such that the new sentence can stand alone, self contained and well defined without the original document context. As not all sentences can be easily edited to stand alone, imagine a sentence from a poetry, models should recognize sentences where decontextualization is not feasible.} 

Many applications of natural language processing need to be able to interpret, or present, text independently from the rich context in which it occurs.
For example, summarization systems extract salient information from documents and present it in a reduced context.
Many systems also segment documents prior to interpretation of retrieval for computational efficiency.
In all of these cases, we would like the context-reduction step to be {\it meaning preserving} but, to date, there has been no independent method of ensuring this.

In this paper we isolate and define the problem of {\em sentence decontextualization}: taking a sentence together with its context and rewriting it to be interpretable out of context if feasible, while preserving its meaning.\footnote{More precisely the truth-conditional meaning or {\em explicature} \cite{sperberwilsonbook}; see section~\ref{sec:linguistic} for discussion.}
Having defined the problem, we operationalize this definition into a high quality annotation procedure; use the resulting data to train models to automatically decontextualize sentences; and present preliminary results that show the value of automatic decontextualization in a user facing task, and as preprocessing for systems that perform document understanding.
We argue that decontextualization is an important sub-task in many downstream applications, and we believe this work can benefit tasks that operate on sentences that occur in a wider context.

\addedcontent{One contribution of this work is to release a dataset of decontextualized sentences that can be used as training and evaluation data, together with the evaluation script: on publication of this paper the data will be available at \url{https://github.com/google-research/language/tree/master/language/decontext}.}

\begin{figure}
\examplebox{
\textbf{Page Title:} Croatia at the FIFA World Cup \\
\textbf{Paragraph:} Croatia national football team have appeared in the FIFA World Cup on five occasions (in 1998, 2002, 2006, 2014 and 2018) since gaining independence in 1991. Before that, from 1930 to 1990 Croatia was part of Yugoslavia. \spansent{Their best result thus far was reaching the 2018 final, where they lost 4-2 to France.} \\
\textbf{Decontextualized Sentence:} 
 The Croatia national football team's best result thus far in the FIFA World Cup
 was reaching the 2018 final, where they lost 4-2 to France.}
\caption{An example decontextualization. The sentence to decontextualize is highlighted in grey.
}
\label{fig:intro_figure}
\end{figure}

Figure~\ref{fig:intro_figure} shows an example decontextualization.
In this example we have a coreference resolution step (their $\rightarrow$ The Croatia national football team's) and a bridging step (insertion of the prepositional phrase ``in the FIFA World Cup" to modify ``Croatia's best result thus far").
Decontextualization involves various linguistic phenomena, including coreference resolution, global scoping, and bridging anaphora~\cite{clark1975bridging}.
We present a linguistically motivated definition of decontextualiation in Section~\ref{sec:linguistic} and show that this definition can be reliably applied by crowdworkers in Section~\ref{sec:task}. 

We generate a corpus of decontextualized sentences corresponding to original sentences drawn from the English Wikipedia.
We show that a high proportion of these original sentences can be decontextualized using a relatively simple set of re-write operations, and we use the data to define a new {\it automatic decontextualization} task in which a computer system needs to create a decontextualized sentence from an original sentence presented in paragraph context. We discuss the implications of choosing Wikipedia as a domain in Section~\ref{subsec:limit}.

We present two methods for automatic decontextualization based on state-of-the-art coreference \cite{Joshi2020SpanBERTIP} and generation \cite{t5} models.
We evaluate the output of these models with automatic measures (derived from ~\citet{Xu2016OptimizingSM}), as well as through human evaluation. 
Both automatic and human evaluations show that the largest sequence-to-sequence model produces high quality decontextualizations in the majority of cases, although it still lags human performance in the thoroughness and accuracy of these decontextualization edits.

Finally, we present two demonstrations of the utility of decontextualization.
The first is a user study giving evidence that decontextualized sentences can be valuable when presented to users as answers in a question-answering task---raters judge that they balance conciseness with informativeness.
In the second one, we use decontextualization as a preprocessing component for generating a retrieval corpus for open domain question answering. Decontextualizing the sentences to be indexed by retrieval system enables more efficient answer string retrieval for information seeking queries.
These demonstrations are presented as preliminary results, and we argue that decontextualization is an important sub-task for a wide range of NLP applications. 

%% file: task.tex
\section{Linguistic Background}\label{sec:linguistic}
We start with the following definition:
\begin{definition}[Decontextualization]
Given a sentence-context pair $(s, c)$, a	sentence $s'$ is a valid {\em decontextualization}
of $s$ if: (1) the sentence $s'$ is interpretable in the empty
context; and (2) the truth-conditional meaning of $s'$ in the empty
context is the same as the truth-conditional meaning of $s$ in context
$c$.
\label{def:decon}
\end{definition}

A context $c$ is a sequence of sentences preceding $s$, and the empty
context is the empty sequence.

We have been careful here to use the more specific term "truth conditional meaning" rather than 
"meaning". Here we follow the distinction in semantics/pragmatics between truth conditional meaning and implicature, and deliberately exclude implicatures (which can also be considered part of the meaning of an utterance) from our definition. 
There is a rich history of work in semantics and pragmatics on truth-conditional meaning and implicatures, going back to \citet{grice1975logic}. 
Our concept of "truth conditional meaning" is very close to "explicature" as used in Relevance Theory \cite{sperberwilsonbook}. Consider this description of explicature from \citet{birner2012} (pages 96--97, our own emphasis added):

``The explicature in an utterance is the result of enriching the
semantic content with the sorts of pragmatic information necessary to
provide us with a truth-evaluable proposition. This includes
calculating the referents for pronouns, working out the intended
interpretation for deictic phrases like here and later ..., \textit{disambiguating lexically and structurally ambiguous words and
phrases, making any ``bridging" inferences necessary for reference
resolution} ... and so on." We will see in the next section that our annotation task follows this definition quite closely. 

As an example consider the following exchange:

\begin{quote}
Susan: Has the Croatia national football team ever won the FIFA World Cup?\\
Jon: Their best result thus far was reaching the 2018 final, where they lost 4-2 to France.    
\end{quote}

Here the truth conditional meaning of Jon's reply is equivalent to "Croatia's best result thus far in the FIFA World Cup was reaching the 2018 final, where they lost 4-2 to France.", whereas the implicature would be "the Croatia national football team has never won the FIFA World Cup" (which answers Susan's question). In our definition the decontextualized sentence $s'$ should preserve the truth-conditional meaning, but is not required to preserve the implicature(s) of the sentence.\footnote{We have not necessarily given up on recovering implicatures: the decontextualized sentence will likely be a valuable intermediate step in deriving the implicatures of an utterance.}

\addedcontent{{\em Remark (extra-linguistic context):} In addition to its document context, a given sentence $s$ and its counterpart $s'$ also come with a temporal, cultural, and geographic context --- i.e. where and when they are being written or read and by whom.\footnote{Research on text simplification~\cite{Xu-EtAl:2015:TACL,bingel-etal-2018-lexi} also shows how target output depends on the expected audience.} We assume that these aspects of context are preserved during decontextualization. The effect of this is that elements of $s$ which derive their meaning from outside of the document context will receive equivalent interpretation in $s'$, and hence do not require decontextualization. For example, the expression "thus far" in Figure~\ref{fig:intro_figure} is interpreted relative to the time of utterance, not relative to what has been previously said in the Wikipedia article, and hence it appears in both the original and decontextualized sentences.}


\section{Task Definition}\label{sec:task}

An annotator is provided with an entire document $d$ with a target sentence within the document, represented as a start and end index $s_{st}, s_{end}$. First, the annotator decides whether the target sentence can be decontextualized or not, labeling it as $\textsc{feasible}$ or $\textsc{infeasible}$. 
If the example is marked as $\textsc{feasible}$, the annotator decontextualizes the sentence, producing $y$, a new sentence that satisfies the conditions in definition~\ref{def:decon}.

\subsection{Feasibility}\label{sec:category}
Sentences in $\textsc{feasible}$ include sentences that do not require any modification to be decontextualized (e.g., \textit{``Émilie du Châtelet proposed 
the hypothesis of the conservation of total energy, as distinct from momentum."}), and sentences that require edits to stand alone.

In the decontextualization step, we instructed annotators to make only minor modifications, which includes copying and pasting a few phrases from the document to the target sentence and deleting phrases from the target sentence. When it is too challenging to decontextualize, it is classified into the $\textsc{infeasible}$ category. Often, sentences in this category are a part of a narrative story, or rely heavily on the preceding few sentences. See Figure~\ref{fig:infeasible} for examples.

\begin{figure}
\examplebox{
\textbf{Page Title:} Postage stamps and postal history of India \\
\textbf{Paragraph}: ... 
In the opinion of Geoffrey Clarke , the reformed system was to be maintained `` for the benefit of the people of India and not for the purpose of swelling the revenue.'' The Commissioners voted to abolish the earlier practice of conveying official letters free of postage (``franking''). \spansent{The new system was recommended by the Governor - General , Lord Dalhousie , and adopted by the East India Company 's Court of Directors.}}
\examplebox{
\textbf{Page Title:} Thermodynamic temperature \\

\textbf{Paragraph}: ...
To completely melt ice at 0 C into water at 0 C, one must add roughly 80 times the thermal energy as is required to increase the temperature of the same mass of liquid water by one degree Celsius. \spansent{The metals' ratios are even greater, typically in the range of 400 to 1200 times.}}
\caption{Decontextualization examples falling into the $\textsc{infeasible}$ category. The sentence to be decontextualized is highlighted in grey.}
\label{fig:infeasible}
\end{figure}

\begin{table*}
\small
\begin{tabular}{p{1.04in}|p{1.3in}|p{3in}|r}
\toprule
\footnotesize{Edit Type}&\footnotesize{Description} & \footnotesize{Example} & \footnotesize{\%} \\\midrule

 {\footnotesize \catt{Pronoun/NP Swap}}
 &{\footnotesize Replacement of a definite pronoun / noun phrase  with another referring expression}
 &{\footnotesize \swaponeone{The copper statue}{The Statue of Liberty}, a gift from the people of France to the people of the United States, was designed by French sculptor Frédéric Auguste Bartholdi and built by Gustave Eiffel.} & 40.5\\ \hline
 {\footnotesize \catt{Name Completion}} &{\footnotesize Expansion of acronyms or partial names}
 &{\footnotesize
  \swaponeone{Meg}{Megan "Meg" Griffin} made her first appearance on television when Family Guy debuted on Fox on January 31, 1999, with the episode ``Death Has a Shadow''.}  & 11.5\\  \hline \hline
 {\footnotesize \catt{DM Removal}}
 &{\footnotesize Removal of discourse markers that can be only understood in context}
 & {\footnotesize \deleteonly{
 For instance,} Alaska could be regarded as the highest state because Denali, at 20,310 feet, is the highest point in the US.
 }& 3.5 \\ \hline
 \hline
 {\footnotesize \catt{Bridging}}&
 {\footnotesize Addition of a modifier (typically a PP) to a noun phrase}
 &
 {\footnotesize In all fights \addonly{in the Ultimate Fighting Championship}, each round can be no longer than five minutes.}&13\\
 \hline
{\footnotesize \catt{Global Scoping}} &{\footnotesize Addition of a phrase (typically a PP) that modifies the entire sentence}
 &{\footnotesize The Japanese film Shoplifters, directed by Hirokazu Kore-eda, won the Palme d'Or \addonly{at the 2018 Cannes Film Festival.}} & 7\\ \hline
 \hline
 {\footnotesize \catt{Addition}} 
 &{\footnotesize Addition of background information that is not necessary but helps readability significantly}
 &{\footnotesize
 Charles Darwin\addonly{, an English naturalist and biologist,} was among the first to suggest that physiological changes caused by an emotion had a direct impact on , rather than being just the consequence of that emotion.} & 10 \\
\bottomrule
\end{tabular}\vspace{-0.3em}
\caption{The list of possible edits in decontextualization. The last column represents how frequently the phenomena occurs in the data, from manual analysis on 200 examples, including examples that belongs to \textsc{infeasible} categories and examples that does not require any edits. The bag notation removes x and adds y ($\lbag$\colorbox{red!20}{-x},\colorbox{green!20}{+y}$\rbag$) at its position.}
\label{tab:edits}
\end{table*}

\subsection{Edit Types and Linguistic Phenomena}
\label{sec:edit_types}

When an example is classified as $\textsc{feasible}$, the annotator makes edits to decontextualize the sentence. Table~\ref{tab:edits} shows the different edit types. They fall into four broad categories:

\paragraph{\catt{Name Completion}, \catt{Pronoun / NP Swap}} correspond to replacement of a referring expression that is unclear out of context with a referring expression that is unambiguous out of context. For example replacing the pronoun ``She'' with ``Cynthia Nixon'', the definite NP ``the copper statue'' with ``The Statue of Liberty'', or the abbreviated name ``Meg'' with ``Megan "Meg" Griffin''.

\paragraph{\catt{DM Removal}} involves removal of discourse markers (DMs) such as ``therefore''.

\paragraph{\catt{Bridging}, \catt{Global Scoping}} involve addition of a phrase (typically a prepositional phrase) that modifies either a particular noun phrase (``bridging'') or the entire sentence (``global scoping''). For example adding ``in the Ultimate Fighting Championship'' as a modifier to ``all fights'', or adding ``at the 2018 Cannes Film Festival" at the end of the sentence. The additional phrase essentially spells out a modifier that is implied by the context. 

\paragraph{\catt{Addition}} inserts background information that significantly improves readability: in many cases, this involves adding an appositive or premodifier to a named entity to add useful background information about that entity. Unlike other edits described above, edits in this category are optional. For example, replacing ``The Eagles'' with ``The American rock band The Eagles.'' 
\input{example_table.tex}

\subsection{Variability} \label{subsec:subjectivity}
We note that for a given sentence frequently there will be more than one possible decontextualization. While this inherent subjectivity makes task challenging to crowdsource and evaluate, we argue this is important feature, as shown in recent literature~\cite{Aroyo2015TruthIA,Pavlick2019InherentDI,kwiatkowski2019natural}, and propose to collect multiple references per example. Table~\ref{tab:diversity} shows examples where there can be multiple different correct decontextualizations. In the first example, while the semantics of the edits are roughly equivalent, i.e., the annotators agreed on what noun phrases have to be disambiguated and information has to be added, they differ in \textit{how} to rewrite the sentence. In the second example, we see disagreement on \textit{what} information should be added to the sentence. We do not make any explicit assumptions about what is known and salient to the reader, and instructed them to use their best judgement to rewrite such that the new sentence is fluent, unambiguous and clear when posed alone. In the last example, annotators disagree on the feasibility. While the sentence is a part of a bigger narrative, two annotators judged it could be edited to alone, by adding a global scoping modifier, ``In Greek mythology." 

\addedcontent{
\subsection{Scope of Current Task Formulation}\label{subsec:limit}
Our data comes from the English portion of the Wikipedia corpus. We sampled sentences as follows. We first pick a (question, Wikipedia, short answer) triple from the Natural Questions~\cite{kwiatkowski2019natural} uniformly at random from the questions that have a short answer. We include the sentence containing the short answer as one example; as a second example we choose a sentence at random from the Wikipedia page. After sampling we exclude (1) sentences under a  ``Plot'' category as they are often infeasible to decontextualize; (2) any sentence that is the first sentence of the page; and (3) any sentence from a paragraph containing only a single sentence. 

We designed this data selection process  to ensure that a large proportion of examples (90\%) could be decontextualized using simple edits described in Section~\ref{sec:edit_types}. 

Before settling on Wikipedia, we conducted an initial pilot study which revealed that encyclopedic text is substantially easier to decontextualize compared to newswire or literary text. In the latter genres, the context required for the comprehension of any given sentence appears to be much more complex in structure. Similarly, it is difficult to posit decontextualization for sentences that appear on social media platforms, because they are situated within complex and highly specific social contexts. In contrast, being written for a general audience, Wikipedia makes limited assumptions about its reader.

Within Wikipedia, we similarly found that articles on popular historical or cultural entities and events were easier to decontextualize by crowdworkers compared to articles from technical domains, such as ones on medical or mathematical concepts. Comprehension of such articles requires a considerable body of background knowledge or information from preceding paragraphs. Articles in our dataset cover topics that require little background knowledge to comprehend.

We focus on decontextualization of \textit{sentences}, where the space of edits is restricted, to make the task easier to quality control and annotate. However alternate formulations, such as decontextualization on \textit{paragraphs} could also be studied. One could even also consider allowing wider range of edits, such as multi-sentence outputs and edits beyond copy-and-pasting, such as paraphrasing and re-ordering. We anticipate exploring such alternative formulations would help to extend the scope of decontextualization to the more challenging domains previously mentioned.
}

We stress however that in spite of our restriction to single sentences in Wikipedia, the decontextualization task is nevertheless valuable: Wikipedia (and other encyclopedic sources) contain a wealth of factual information, and a high proportion (over 60\%; see table~\ref{tab:data_stat}) of sentences both require decontextualization and can be decontextualized under our definitions (only 30\% of sentences are interpretable out of context without any edits).

%% file: example_table.tex
\begin{table*}[t]
\footnotesize
\centering
\begin{tabular}{p{\textwidth}} \toprule
\textbf{Page title / Section title}: We Don't Talk Anymore (Charlie Puth song) / Music video \\
\textbf{Paragraph}:
The music video premiered on August 2 , 2016 , on BuzzFeed and was directed by Phil Pinto . \spansent{It shows Puth and Mirella Cardoso as his love interest}. $\ldots$ \\
\textbf{Decontextualization 1:}  $\lbag \colorbox{red!20}{{-It}},
       \colorbox{green!20}{{+We Don't Talk Anymore music video}} \rbag$ shows $\lbag \colorbox{red!20}{{-Puth}},
       \colorbox{green!20}{{+Charlie Puth}}\rbag$ and Mirella Cardoso as his love interest.\\
\textbf{{Decontextualization 2}:}  $\lbag \colorbox{red!20}{{-It}},
       \colorbox{green!20}{{+The ``We Don't Talk Anymore''(Charlie Puth song) music video}}\rbag$  shows Puth and Mirella Cardoso as his love interest.\\
\midrule
\textbf{Page title}: The American Baking Competition \\
\textbf{Paragraph}: CBS placed casting calls for participants on November 14, 2012 . Auditions were held between December 1 and December 15, 2012. \spansent{The competition took place at the Gibbs Gardens in Ball Ground , Georgia in March 2013.}\\
\textbf{Decontextualization 1}: The \swaponeone{competition}{American Baking Competition} took place at the Gibbs Gardens in Ball Ground , Georgia in March 2013.\\
\textbf{Decontextualization 2}: The \swaponeone{competition}{American Baking Competition, a reality competition television series,} took place at the Gibbs Gardens in Ball Ground , Georgia in March 2013 .\\
\midrule

\textbf{Page title}: Gemini (Constellation) \\
\textbf{Paragraph}: In Greek mythology, Gemini was associated with the myth of Castor and Pollux, the children of Leda and Argonauts both. Pollux was the son of Zeus, who seduced Leda, while Castor was the son of Tyndareus, king of Sparta and Leda's husband. Castor and Pollux were also mythologically associated with St. Elmo's fire in their role as the protectors of sailors. \spansent{When Castor died, because he was mortal, Pollux begged his father Zeus to give Castor immortality, and he did, by uniting them together in the heavens.}\\
\textbf{Decontextualization 1}: \textsc{Infeasible}\\
\textbf{Decontextualization 2}: \addonly{In Greek mythology,} when Castor died, because he was mortal, Pollux begged his father Zeus to give Castor immortality, and he did, by uniting them together in the heavens. \\
\bottomrule

\end{tabular}\vspace{-0.5em}
\caption{Examples showing the diversity of valid decontextualization edits.}\label{tab:diversity}
\end{table*}

%% file: data.tex
\section{Data Collection} \label{sec:data}

\paragraph{Annotation Interface}

The annotator is presented a sentence in the context of an entire Wikipedia page. In the first step the annotator judges whether the example is \catt{feasible} or \catt{infeasible}. If the example is marked as \catt{feasible}, the annotator can use delete, add, or swap operations within a user interface to produce a decontextualized string. 
\commentout{
We have designed a three-stage annotation process, where the annotator is to (1) first decide whether sentence makes sense on its own, (2) decide whether decontextualization is feasible, and (3) perform decontextualization through one or many of the following edit operations. We constrained the space of possible edits to following four categories: \textsc{Deletion}, \textsc{Name Swap}, \textsc{Noun Phrase Swap}, and \textsc{Addition}. 

The description of each operation follows:
\begin{itemize}
\setlength{\itemsep}{0pt}%
    \setlength{\parskip}{0pt}%
    \item  \textsc{Deletion} annotators will identify discourse connectives (e.g., ``however", ``also") that require understanding of a large chunk of previous sentences
    \item \textsc{Name Swap} identify pronouns (e.g. she, he, it, I, they, her, its, etc), acronyms (e.g., MPAA) or partial names (e.g., Lee, Thomas) in the sentence and specify them fully. 
    \item \textsc{NP Swap} identify nouns (e.g., the bridge, this) that are ambiguous/underspecified when taken out of context. 
    \item \textsc{Addition} add tokens to a sentence to add contextual information (e.g., to disambiguate when the events occurred).
\end{itemize}}

\begin{table}
\scriptsize
    \centering
    \begin{tabular}{l|r|rr|rr|r }
\toprule
  & \#  & par.    & sent.  &\multicolumn{2}{c|}{$\textsc{feasible} (\%)$} & $\textsc{infea}$\\
  & & len & len  & w/ edit & as is & \textsc{sible }(\%)\\ \midrule
Train & 11290 & 695	&156 &	60&	31&	9\\
Dev & 1945 & 695	&	162& 67 & 21&12 \\
Test  & 1945 & 711&		160 &68&20 & 12 \\
\midrule
Expert & 100 & 658 & 163 &  63 &26 & 12 \\
 \bottomrule
    \end{tabular}\vspace{-0.3em}
    \caption{Data statistics. par. len refers to paragraph length in bytes, and sent. len refers to sentence length in bytes. All lengths are in bytes. The development and test set is five-way annotated, and the expert data is four-way annotated.}
    \label{tab:data_stat}
\end{table}
\paragraph{Data Statistics}
We collected one reference for each example in the training data, and five references for each example in the evaluation data. Annotators are native speakers of English located in the United States, and on average, they took 4 minutes to annotate a single example. 

In total, 28 annotators annotated the examples, with 11 annotators annotating more than 1K examples each. 

Table~\ref{tab:data_stat} represents some overall data statistics. Decontextualization is possible for the majority of examples, with the $\textsc{infeasible}$ category covering roughly ~10\% of the data. We note a slight discrepancy between train and evaluation dataset distribution, potentially due to a change in the annotation interface. A small subset of data is annotated by the authors to be compared with the crowd-sourced data (last row in the table). 
\paragraph{Annotation Quality}
We quantify the annotation agreement on the category classification. The Fleiss' kappa on category classification is 0.51 among expert annotators, and is 0.30 among the crowd annotators (binary agreement is at 85\%). We observed more variability in crowdworkers as annotators' background is more diverse, and some annotators have a loose concept of ``stand alone" and consistently attempted decontextualization. 

We also measured agreement among the individual edits. For each of the edit operations (as defined in Section~\ref{sec:edit_types}), we compare the output sentence after the single edit and to a set of output sentences, each after a single edit by other annotators. About 32.5\% of edits were covered.

Because of the inherent annotation variability, four of the authors manually evaluated 100 crowd-sourced annotations from the evaluation data based on two measures: (1) whether the sentence is sufficiently and correctly decontextualized, and (2) whether the sentence is grammatically correct and fluent. Overall, 88\% of annotations were valid in both, 89\% on the content and 88\% on form. 

%% file: evaluation.tex
\section{Automatic Decontextualization}\label{sec:automatic_decontextualiztion}
\subsection{Models}
We present two models for decontextualization: a coreference resolution model and a sequence-to-sequence generation model.
For both models, the input is a concatenation of the title of the Wikipedia document, the section titles, and the paragraph containing the target sentence. During the annotation pilots, we found the document title is crucial for decontextualization, while section headers were frequently necessary or missing. We denote the title of the Wikipedia page as the sequence of tokens $t$, section titles of the paragraph as the sequence of tokens $t_s$ and the $n$ sentences in the paragraph where the target sentence is coming from as $x_1 \ldots x_n$, where each $x_i$ is a sequence of tokens, and $x_t$ is the target sentence ($1\le t \le n$).
The model considers the concatenation of a subset of the document,
\begin{equation*}
\footnotesize
\texttt{[CLS]} t \texttt{[S]}  t_s \texttt{[S]} x_1  \cdots x_{t-1} \\
\texttt{[S]} x_t \texttt{[S]} x_{t+1}\cdots x_{n} \texttt{[S]}
\end{equation*}
where $\texttt{[S]}$ is a separator token. This representation differs from the setting of annotators, where they were given the full document context. \addedcontent{
As an approximation, we include article and section titles in the inputs, as these often contain salient contextual elements.}
We did experiment with giving more context, i.e., adding the first paragraph of the article as an additional input, but did not observe a performance improvement. On the initial pilot, annotators marked that 10-20\% of examples required access to the full document.

\paragraph{The Coreference Model}
As many decontextualization edits can be recovered by a coreference resolution module, we adapt the output from the state-of-the-art coreference resolution system of~\cite{Joshi2020SpanBERTIP}, trained on the CoNLL dataset~\cite{Pradhan2012CoNLL2012ST}, as a decontextualization system. We used the publicly available pre-trained checkpoint of SpanBERT-Large with the original hyper parameters.\footnote{https://github.com/facebookresearch/SpanBERT/} 

We run this model on the input sequence, and map the coreference cluster predictions to modify the sentence as follows. We only consider clusters with a mention in the target sentence. For each such cluster, we find its first mention inside the target sentence, and find another mention in the same cluster that was presented earlier in the input and is longer than the current mention. If such a mention is found, we replace the current entity mention string with the earliest such mention string (e.g., "She" is replaced with "Taylor Swift"). On average, 36.5\% of examples were modified through this process.

\paragraph{The Seq2Seq Generation Model} is based on the recent T5 model~\cite{Raffel2019ExploringTL}.
We show two variations of the model, BASE and 11B, which mainly differ in the model capacity. We fine-tune the model on our crowdsourced training set, by setting the target sequence to be $ \texttt{[CAT]} \texttt{[SEP]} y$, where $\texttt{[CAT]}\in \{\textsc{unnecessary}, \textsc{feasible}, \textsc{infeasible}\}$ and 
$y$ is a decontextualized sentence when $\texttt{[CAT]} = \textsc{feasible}$ and the original sentence when $\texttt{[CAT]}\in \{\textsc{unnecessary}, \textsc{infeasible}\}$. \textsc{unnecessary} are examples where the original sentence without any edit can stand alone.

We limit the input/output to 512/128 tokens for both variants, and fine-tuned from pre-trained checkpoints\footnote{https://github.com/google-research/text-to-text-transfer-transformer} with a batch size of 100 examples until the validation loss stopped decreasing, after about 32K for the larger and 500K steps for the smaller model.

\subsection{Evaluation}

\subsubsection{Feasibility Detection}
We first evaluate the accuracy of models in making the feasible vs. infeasible decision.
To do this we compute the binary agreements with all human references and average them to get an accuracy. 

\paragraph{Results}
For the feasible vs. infeasible classification task, baseline that always predicts \textsc{feasible} will have 88\% accuracy. The larger variant of T5, T5-11B, achieves 89\% accuracy, outperforming human agreement (85\% accuracy), affirming the strong performance of pre-trained language models on classification tasks~\cite{bert}. This model predicts the \textsc{infeasible} category infrequently for the larger variant (5\% of examples), while humans classify an example as \textsc{infeasible} for 12\% of examples. We observe the smaller variant, T5-Base, is less accurate, over-predicting the \textsc{infeasible} category (for 20\% of examples), getting 77\% accuracy. The coreference model cannot decide the decontextualization feasibility, as an untrained baseline.

\commentout{Mike: This seems redundant, we can mention this when we give this result. We also report how frequently model predicted infrequent label, \textsc{infeasible}, as always predicting \textsc{feasible} will score high with the accuracy metric.}

\subsubsection{Decontextualized Sentence Generation}

\paragraph{Setup}
For development / test examples, we have five human annotations per example. We only consider examples marked by three or more annotators (out of five) as \textsc{feasible} for decontextualized sentence generation. For each of these examples, we discard annotations which mark the example as \textsc{infeasible}. For automatic evaluation and comparison, we need a \textcolor{blue}{human output}, which will be compared to model outputs, and a set of \textcolor{purple}{reference annotations} which will be considered as correct, gold annotations. The single human output provides a reference point for evaluation measures to which the automatic output can be compared.

We observed comparing a longer decontextualized sentence to shorter decontextualized sentences often erroneously results in low scores automatic metrics (e.g., In the last example of Table~\ref{tab:diversity}, adding extra information will be erroneously punished). Thus, instead of randomly selecting one annotation to be used as the representative \textcolor{blue}{human output}, we sort the annotations by the length of the output sentence (raw bytes), and take the annotation with median length\footnote{When there are four of references, we take the second shortest sentence.} as a \textcolor{blue}{human output} and take the remaining annotations as a set of \textcolor{purple}{reference annotations}. From manual inspection of the data the median-length output appeared often to be optimal in terms of balancing length versus accuracy of the decontextualization.

\paragraph{Metric}
For each model prediction and \textcolor{blue}{human output}, we report:

\begin{itemize}[leftmargin=0.15in]
\setlength\itemsep{0.1em}
  \item Length Increase, the average value of (len(decontext)-len(original)) / len(original).
  \item \% edited, the proportion of examples that were modified for decontextualization (as opposed to being left unchanged).
  \item Sentence match, a binary score computed between the output and a set of references, indicating whether the output matches any of the \textcolor{purple}{references} after normalization (stripping away articles and punctuation and lowercasing). We report two numbers, a score on all examples, and a score on examples where all references edited the sentence.
  \item SARI (\textbf{s}ystem output \textbf{a}gainst \textbf{r}eferences and against the \textbf{i}nput sentence) metric~\cite{Xu2016OptimizingSM}. To compute this, for each \textcolor{purple}{reference}, we calculate a set of \textbf{add} edits, corresponding to which unigrams are seen in the \textcolor{purple}{reference} but not in the original sentence. Conversely, we can calculate the set of \textbf{delete} edits, corresponding to unigrams that are in the original sentence but not in the \textcolor{purple}{reference}. We calculate precision/recall/F1-measure on add and delete edits. We look at unigrams only, and use fractional counts for the words in the \textcolor{purple}{references} (i.e., a word appearing in one of $r$ references will be counted as 1/$r$). We compute micro average across examples, i.e., globally by counting the total true positives, false negatives and false positives, as many examples do not require any edits.\footnote{Similar to BLEU in machine translation, SARI is a useful measure for comparing different systems; however, due to the relatively large space of possible decontextualizations it will not be possible to achieve anything close to 100\% F1 on SARI measures, and thus the absolute score is harder to interpret. A SARI score of for example 50\% should {\em not} be interpreted as indicating a system with 50\% accuracy.} 
\end{itemize}
While the sentence match score is the easiest to interpret, it punishes longer outputs, making comparisons across systems producing outputs of different lengths challenging, and it overly rewards conservative strategies that simply copy across the original sentence. Thus, we use the SARI metric as our main evaluation metric. SARI can be thought of as a precision/recall measure on topics (unigrams) that should be added or deleted. 
\setlength{\tabcolsep}{4.5pt}
\begin{table}
\footnotesize
    \centering
    \begin{tabular}{l|rr|r|cc}
\toprule
   & len &\% ed&{match} &  SARI add&  {SARI del}\\
   & inc.&  ited & all /  edited& F1 (P/R)&F1 (P/R)  \\ \midrule
{Repeat} & 0& 0 & 38  / 0 &\fpr{0}{0}{0}  &\fpr{0}{0}{0} \\
Coref  & 7&42&39 / 13&\fpr{22}{51}{14}&\fpr{31}{34}{28} \\ 
T5-Base&8&40 & 48 / 21&\fpr{29}{67}{19}&\fpr{40}{54}{32}  \\
T5-11B&      12&59&53 / 32&\fpr{42}{72}{30}&\fpr{46}{49}{43}  \\
\textcolor{blue}{Human} &   24&76&45 / 29&\fpr{56}{64}{49}&\fpr{58}{61}{55}    \\
\bottomrule
    \end{tabular}
    \caption{Development set performance. Len inc. is the average percentage increase in length from decontextualization. \% edited is the proportion of examples that have at least one edit. match-all shows percentage of outputs that have at least one match in the human references; match-edited shows the match value calculated on cases where all references include at least one edit.}
    \label{tab:dev}
\end{table}

\begin{table}
\footnotesize
    \centering
    \begin{tabular}{l|rr|r|cc}
\toprule
  & len &\% ed&{match} &  SARI add& SARI {del}\\
  & inc.&  ited & all / edited & F1 (P/R)&F1 (P/R)  \\ \midrule
Repeat &0 & 0 &  36 / 0 &\fpr{0}{0}{0}  &\fpr{0}{0}{0}\\
Coref &8 & 42& 38 / 13& \fpr{23}{50}{15} & \fpr{36}{40}{32}\\
T5-11B&  13&61&52 / 32&\fpr{43}{69}{31}&\fpr{47}{49}{46}\\
\textcolor{blue}{Human} & 23&77&44 / 28  &\fpr{56}{64}{49}&\fpr{58}{61}{56} \\
\bottomrule
    \end{tabular}
    \caption{Test set results. See table~\ref{tab:dev} caption for a key.}
    \label{tab:test}
\end{table}

\paragraph{Automatic Evaluation}

Tables~\ref{tab:dev} and~\ref{tab:test} show development and test performance. A successful decontextualization system would result in high sentence match, adequate changed ratio (experts edited about 79\% of examples), and length change ratio (the experts' ratio is 1.19), as well as high SARI addition and deletion scores. As a sanity check, we report \textsc{Repeat}, which outputs the original sentence. This alone results in high sentence match score, around 40\%, meaning that on this number of examples, at least one of the annotators deemed the sentence can stand alone without any edits.

The coreference system has an exact match of about 13\% of examples that require edits, without any task-specific fine-tuning. Its SARI add scores shows high precision and low recall, and its deletion scores are low as it cannot delete discourse markers. The Seq2seq generation model achieves high scores across all measures. The bigger variant is substantially better, editing more than its smaller variant without losing precision. We observe the larger variants outperform the average human on sentence match measure, but not in SARI measures. The T5 model modifies fewer examples than the annotator, and edits involve fewer tokens, benefiting it on the sentence match measure. However, the model is more likely to miss required edits, as shown in low recall for the SARI add and deletion measures. We discuss this further in the following human evaluation section.

\begin{table}
\footnotesize
    \centering
    \begin{tabular}{l|rrr|r}
\toprule
  & T5 & either & Annotator & Sum  \\ \midrule
T5 & 13 & 12 & 2 & 27\\
either & 7&22&4  & 33 \\
Annotator & 1&15&24  & 40\\ \midrule
Sum & 21 & 49 & 30 &100\\
 \bottomrule
    \end{tabular}
    \caption{Preference between T5 output and human annotation. Columns represents the judgement of the expert A, rows that of the expert B. We see high agreement between two expert annotators, despite one expert annotator (column annotator) is ambivalent more frequently.}
    \label{tab:human_agreement}
\end{table}

\paragraph{Human Evaluation}
We sampled 100 examples in the evaluation set, where at least two annotators \textbf{and} our best model made decontextualization edits. We randomized the order or presentation of the T5 and human outputs so as to not  bias the annotation. On this set, we (two of the authors) conducted a manual evaluation. Given two decontextualized sentences, one from the best model and another randomly selected from a set of annotations with decontextualization edits, we evaluated each on two dimensions: (a) is it fluent and grammatically correct? (b) is it sufficiently and correctly decontextualized? Lastly, we chose the preference between two outputs (A, B, or either). 

Expert annotators marked as `sufficient' those items for which all possible referential ambiguities had been resolved. Given the subjective nature of the task, some `insufficient' decontextualizations by the expert annotator could be valid for the another annotator with a different world knowledge. We report averaged binary scores from two experts. The model output scored 88.0\% on fluency, and 67.5\% on correct decontextualization, while the human reference output scored 84.5\% on fluency and 78.5\% on correct decontextualization. Both annotators found T5 to be slightly more fluent, while humans are more thorough and accurate in decontextualizating. Table~\ref{tab:human_agreement} shows the preferences of two annotators. Both preferred human output, and their preferences exhibit high agreement (matching on 37 out of forty examples when both had preferences). 

We briefly characterize common error patterns for annotators and the T5 model. Similar error patterns emerge between the annotations and model outputs. Both occasionally fail to identify generics that need to be replaced with referring NPs, phrases that require bridging, and temporal contexts that should be provided. Additionally, we noticed that the T5 model heavily relies on the title cues, and sometimes fail to clarify ambiguous entities that are not the main entity of the page. We noticed very few examples where T5 hallucinates factually incorrect contents.

%% file: userstudy.tex
\section{Two Applications}
We present two demonstrations of the utility of decontextualization. First, we argue that the decontextualized sentences can be valuable in themselves in question answering, and show that they can be useful as a preprocessing step.

\begin{table}
\footnotesize
    \centering
    \begin{tabular}{l|rrr|c}
\toprule 
& \multicolumn{3}{c|}{Prefer}& log odds \\
Opt.A vs. Opt.B & A &  B & either  & intercept [CI]\\\midrule
Dec. vs. Ori. & 730  & 426 & 364 &0.85 [0.4,1.3] \\
Dec. vs. Par. & 850 & 456 & 234 &0.55 [0.1,1.0]\\
Ori. vs. Par. & 741 & 505 & 274 &0.31 [-0.2,0.8] \\
 \bottomrule
    \end{tabular}
    \caption{User study results. Dec. refers to decontextualized sentence answer, Ori. means original sentence answer, Par. means paragraph answer. We present raw counts of preferences and the log odds of preferring option A and its 95\% confidence interval.}
    \label{tab:userstudy}
\end{table}
\subsection{Decontextualized Answer As Is}\label{subsec:userstudy}
\begin{figure*}
\centering
{\includegraphics[width=\textwidth]{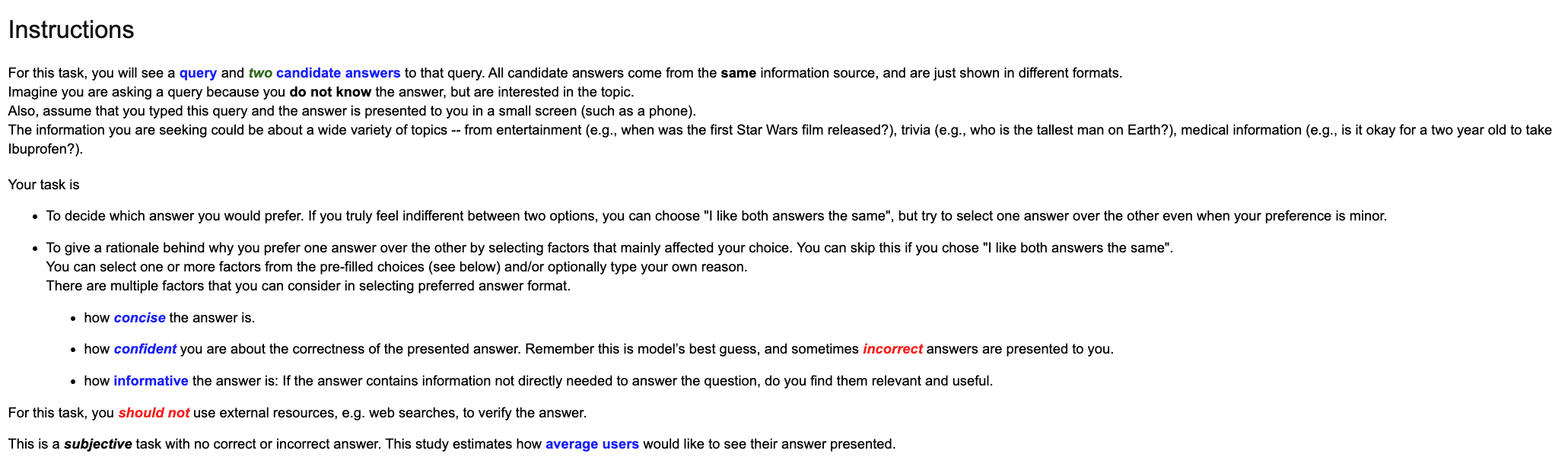}}
\noindent\rule{15.5cm}{0.5pt}

{\includegraphics[width=\textwidth]{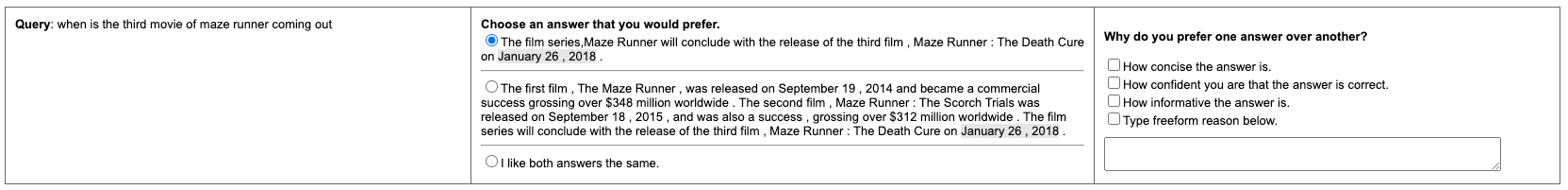}}
\caption{A screenshot of the instruction and an example instance comparing the original sentence answer and the paragraph answer shown to annotators for the user study.}\label{fig:user_study}
\end{figure*}

We showcase a use case of decontextualized sentences as providing a succinct yet informative answer to open domain factoid questions~\cite{kwiatkowski2019natural}. We design a user study where people compare a decontextualized-sentence answer with an original-sentence answer and a paragraph answer to the same query.\footnote{Understanding how to present answers to users is a complex problem with many desiderata, e.g., preserving the original content, crediting the source, interaction with the user interface, which we are not covering comprehensively.}
\paragraph{Setup}
Given a question and two presentations of the same answer, raters were tasked with marking their preference between the two answer presentations (option A, option B, or either). The actual short span answer in the sentence is always highlighted (similar to seen in Table~\ref{tab:user_study_examples}) (See Figure~\ref{fig:user_study} for a screenshot).


We conduct three comparison studies on the same set of 150 questions: (a) decontextualized sentence vs. original sentence, (b)
original sentence vs. original paragraph, (c) decontextualized sentence vs. original paragraph. For each example in each study, we collected 10 user ratings. The questions are randomly chosen from a set of questions which have a short answer, and such that the sentence containing the short answer is categorized as $\textsc{feasible}$ by the annotators and edits were necessary to decontextualize. We use crowd-sourced annotations of decontextualized sentences. Figure~\ref{fig:user_study} shows the screenshot of the user study interface. 

\paragraph{Result}
Table~\ref{tab:userstudy} shows the results of the user study. We observe that decontextualized sentence answers are preferred to both the original sentence answers and the original paragraph answers. We also note that the users preferred sentence answer compared to paragraph answer in general.

\begin{figure}
\centering
\includegraphics[width=0.45\textwidth]{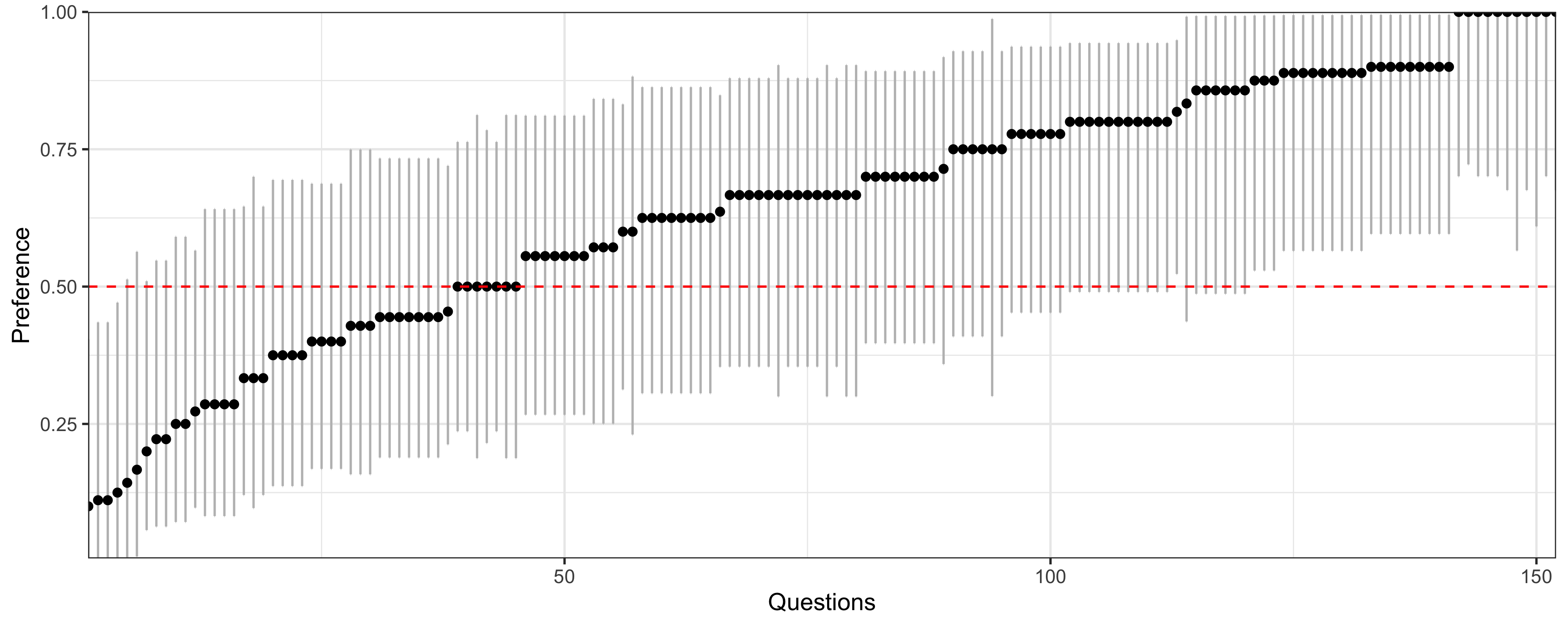}
{\includegraphics[width=0.45\textwidth]{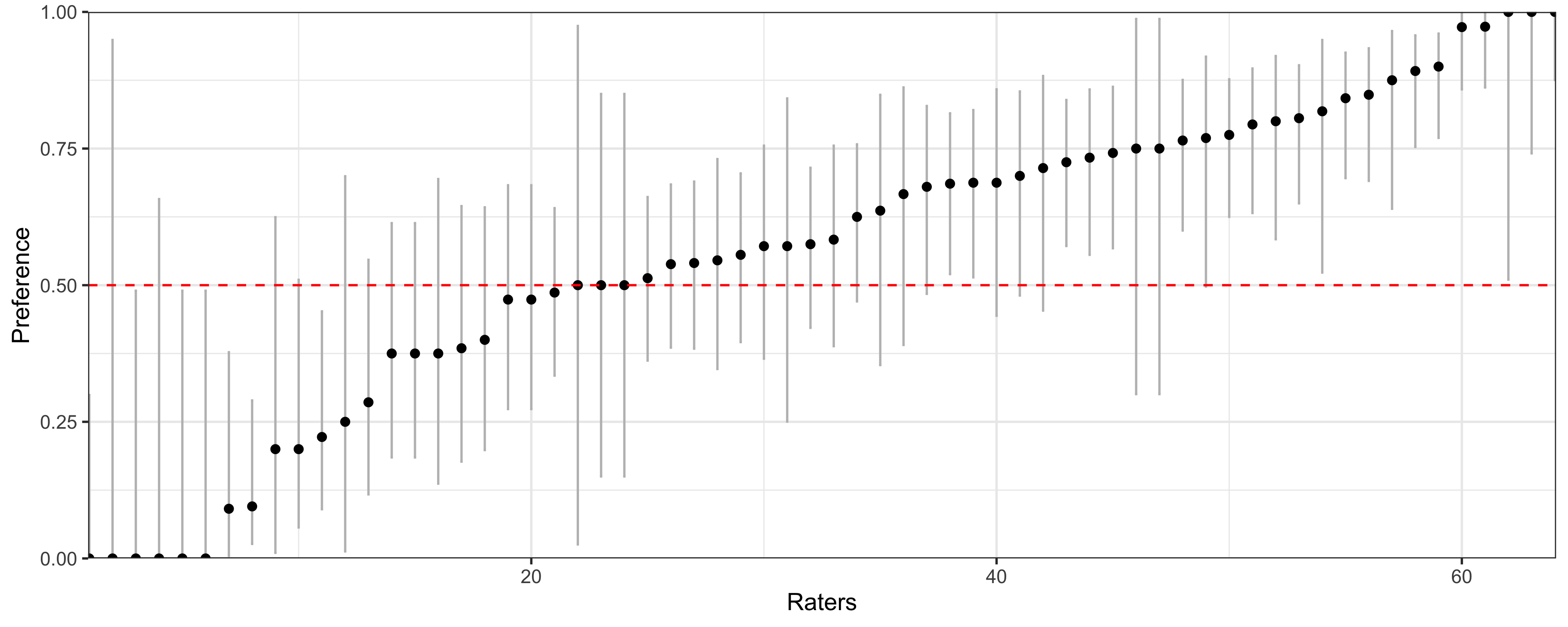}}\vspace{-0.2em}
\caption{Each dot represents how frequently the decontextualized answer is preferred for {a single question} / {rater} to the original sentence for a single question (top plot) and single rater (bottom plot). Questions (top plot) and raters (bottom plot) are sorted by its preference towards the decontextualized answer. The red line is where both are equally preferred, and above the line represents question / rater where decontextualized answers were preferred. While the decontextualized answer is preferred overall, we see a large variability.
}\label{fig:preference_variability}
\end{figure}

We further investigated the statistical significance of the preferences reported in Table~\ref{tab:userstudy}. We noticed a quite large amount of question and rater variability---some raters consistently preferred a sentence answer, valuing conciseness, while some raters behaved in the other direction. Similarly, for some questions, all raters preferred a sentence answer. Figure~\ref{fig:preference_variability} visualizes such variability based on the questions and raters.


\begin{table*}
\footnotesize
    \centering
    \begin{tabular}{p{.8in}|p{1.7in}|p{2.4in}|p{0.3in}|p{0.3in}}
\toprule 
\vspace{-0.5em}
{Query} & {Decontextualized answer} & {Paragraph answer (sentence answer highlighted)} & { Decont.}& {Ori.} \\\midrule
when was the rising of the moon written	&	The Rising of the Moon, Irish ballad recounting a battle between the United Irishmen and the British Army 
has been in circulation since \spanone{circa 1865}.&	\spansent{The ballad has been in circulation since }\spanone{circa 1865}. The earliest verifiable date found in publication is 1867 .&-2.09	&-1.53\\\midrule
what is the most viewed video on youtube in 24 hours&The most viewed music video within 24 hours of its release is \spanone{Taylor Swift 's Look What You Made Me Do}.	&This list of most viewed online videos in the first 24 hours contains the top 30 online videos that received the most views within 24 hours of release across the world. This list excludes movie trailers , which are featured on the list of most viewed online trailers in the first 24 hours. \spansent{The most viewed music video in this time period is }\spanone{Taylor Swift 's Look What You Made Me Do} .&	1.06	&-0.48	\\
 \midrule
 when was last time england got to quarter finals in world cup	&	The England national football team have reached the quarter - finals on nine occasions , the latest of which were at the 2002 and the \spanone{2006}.&	England did not enter the competition until 1950\ldots
 Their best ever performance is winning the Cup in the 1966 
 , 
 whilst they also finished in fourth place in 1990, and in 2018. 
 \spansent{Other than that, the team have reached the quarter - finals on nine occasions, the latest of which were at the 2002 and the} \spanone{2006}.&1.40&	0.70\\
 \bottomrule
    \end{tabular}
    \vspace{-0.5em}
    \caption{Examples from the user study. The last column represent coefficients for preferring original sentence over the original paragraph, and the fourth column presents coefficients for decontextualized sentence over the paragraph. Positive values means preference towards the sentence-length answer over the paragraph-length answer.}
    \vspace{-0.5em}
    \label{tab:user_study_examples}
\end{table*}

To control for the correlations induced by the rater and question groups, we fit a generalized linear mixed model (GLMM) using the \texttt{brm} R package~\cite{brms}. For this analysis, we excluded data points where users did not show a preference (selected either). We used the formula:
$\texttt p \, \mathtt \sim \, \texttt{1 + (1|r) + (1|q)}$
where \texttt{p} is whether a rater chose one option over the other; \texttt{r} is the rater id; and \texttt{q} is the question id. This formula specifies a regression of the log-odds of the rater preference while allowing for random effects in the raters (\texttt{r}) and questions (\texttt{q}). The last column of Table~\ref{tab:userstudy} shows the fixed effect coefficients and their confidence intervals. The intercept represents the strength of preference towards option A. We found a statistically significant preference for decontextualized sentences over both original sentences and the paragraphs (p-value was smaller than 0.05 for both studies).




\paragraph{Examples}
We qualitatively investigated which examples benefit from decontextualization, and in which examples  raters prefer paragraph answers. Table~\ref{tab:user_study_examples} shows questions together with two answer presentations, along with the predicted fixed effect question coefficient towards decontextualized answer in study (b) and towards the sentence answer in study (c). In the first row, the added information from the decontextualization is not relevant to the question, thus we observe preference \textit{against} decontextualization. In the second and third row, the decontextualized sentence answer is preferred as it provides enough evidence to answer the query, while the original sentence answer does not.

%% file: applications.tex
\subsection{Decontextualizing System Inputs}\label{sec:open-qa}
Having shown the benefits of decontextualization in a user facing task, 
we now investigate the use of decontextualizaton as a preprocessing step. Specifically, we construct a passage retrieval corpus for open domain question answering \cite{drqa} with decontextualized sentences. Experiment shows that decontextualized sentences ensure completeness of the passages while minimizing their length (thus computational cost). 

\paragraph{Background}
Open domain question answering typically consists of pair a passage retrieval~\cite{Liu2002PassageRB} and transformer-based answer extractor (reading comprehension model) based on the retrieved passages~\cite{realm,Karpukhin2020-lt,Izacard2020-gt}. 
And the computational cost is dominated by the cost of co-encoding the query with the retrieved passages (typically paragraphs or overlapping 100 word windows). 

\paragraph{Setup}
We create a corpus using the 7k documents (233k paragraphs, 868k sentences) from the documents associated with the questions in the NQ-open development set \cite{orqa}.   
We consider a retrieved passage to be correct if it contains one of the answer strings\footnote{We adopt the answer match heuristics from \citet{orqa}.} and investigate the number of questions for which we can retrieve a correct passage for a fixed computational cost. Under this measure, we compare paragraphs, windows of 100 words, sentences, and decontextualized sentences as a set of retrieval passages. These segmentation approaches generate different number of passages for the same article (paragraph and a window of 100 words segmentation make fewer passages compared to sentences-level segmentation). To generate decontextualized sentences process all paragraphs with T5-11B model, which are trained on all annotated data (including development and test set). For about 40\% of sentences, the model classified the sentence as infeasible to decontextualize or unnecessary to make any edits, we use the original sentence. On the other 60\% model tended to add more information. For example, for a sentence ``Bush was widely seen as a `pragmatic caretaker' president who lacked a unified and compelling long-term theme in his efforts.", the decontextualized sentence will be ``George H.W. Bush was widely seen as a `pragmatic caretaker' president of the United States who lacked a unified and compelling long-term theme in his efforts." and a paragraph would be entire paragraph containing this sentence, and 100-word window will be a chunk without using a sentence boundary as a segmentation. For all, we prepend the document title to the passage, following the literature and use the TFIDF as a retriever model.

\paragraph{Metric}
Let $\mathbf{q_i}$ be a question; let $\mathbf{A}_i = [a_i^0 \dots a_i^n]$ be the set of valid answers; let $\mathbf{C}_i = [\mathbf{c}_i^1 \dots \mathbf{c}_i^k]$ be a ranked list of evidence passages; and let $H(\mathbf{A}_i, \mathbf{C}_i)$ be the index of the top ranked context that contains one of the valid answers (or $k + 1$ if there is no such context).
We first define the cost of encoding a single question and passage, $c(\mathbf{q_i}, \mathbf{c_i^m}) = (|\mathbf{q_i}| + 1 + |\mathbf{c_i^m}|)^2$.
This captures the fact that the Transformer's computation cost scales quadratically with the length of the encoded text (question + separator + evidence passage). 

\begin{align*}
O(\mathbf{q_i}, \mathbf{A_i}, \mathbf{C_i}) &= \sum_{m=1}^{H\left(\mathbf{A}_i, \mathbf{C}_i\right)}   c(\mathbf{q_i}, \mathbf{c_i^m}). 
\end{align*}

Given the per example cost defined above, we define the \textbf{\textit{recall}} of the retrieval system at computational cost budget $t$ to be:
\begin{equation}
{
 \frac{1}{N}\sum_{i=0}^N {\mathbbm{1}}\left[ O(\mathbf{q_j}, \mathbf{a_j}, \mathbf{C_j}) < t \right]
}\label{eqn:retrieval_complexity}
\end{equation}
\noindent where $N$ is the total number of examples and $\mathbbm{1}$ is an indicator function. We use this as an evaluation measure instead of mean reciprocal rank or recall at N, to compare across different retrieval passage length. 
\begin{figure}
    \centering
    \includegraphics[width=0.43\textwidth]{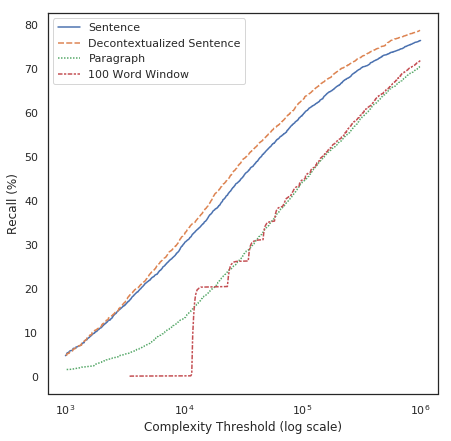}
    \caption{Retrieval recall plotted against computational cost budget (Eqn~\ref{eqn:retrieval_complexity}) for different methods of document segmentation.}
    \label{fig:retrieval_complexity}
\end{figure}
\paragraph{Results}
Figure~\ref{fig:retrieval_complexity} plots the recall of each retrieval corpus at different computational cost budget $t$ on the whole NQ-open evaluation set. 
The graph shows that sentence level segmentation is more cost-effective than paragraph or 100 word level segmentation, and using decontextualized sentences is more cost effective than using the original sentences. Decontextualized sentences near the performance of commonly used 100 word windows with 1/10th the cost.

This result exemplifies the way in which decontextualization can be used to ensure that the input to natural language understanding system is concise yet complete. We think this way of using decontextualization as a preprocessing could also aid tasks such as summarization.

%% file: related.tex
\section{Related Work}
Prior literature in summarization studied how article context affects the understanding of sentences within an article. It has been observed that disambiguating entity mentions and correctly resolving anaphora is crucial for automatic summarization~\cite{Otterbacher2002RevisionsTI, steinberger2007} and for evaluation of summarization systems~\cite{Pitler2010AutomaticEO}. \citet{Li2016ImprovingTA} identified information missing from a sentence could be identified in the article context in newswire text 60\% of the time. This is considerably less frequent than for the encyclopedic text studied here, but nevertheless hints that decontextualization for newswire text could be feasible. It remains unclear whether information accessible in newswire contexts can be readily incorprated into sentences using controlled edits of the type we employ.

Successful decontextualization models must resolve entity and event coreferences~\cite{Humphreys1997EventCF} as well as other forms of anaphora~\cite{Rsiger2018BridgingRT}. These are necessary but insufficient for decontextualization however, which also involves discourse marker removal, acronym expansion, and fluent and grammatical sentence generation.

The term decontextualization was introduced in a recent table-to-text generation dataset~\cite{Parikh2020ToTToAC} where a sentence from a Wikipedia document was \textit{decontextualized} such that it can be interpretable when presented with a table alone. They cover only the sentences that are relevant to the table, and adapt it to the table context. In a recent image captioning dataset~\cite{Sharma2018ConceptualCA}, sentences are re-written such that information that cannot be inferred from the image is removed. For example, entity names are replaced with generics (e.g. \swaponeone{Tom Cruz}{A man} is waiting."). 

%% file: conclusion.tex
\section{Conclusion}
We define \textit{decontextualization}, the task of rewriting a sentence from a document to be interpretable in an empty context, while preserving its meaning. We build a crowdsourced dataset and a model for decontextualization, and demonstrate how decontextualization can be used in a user-facing task and as a sub-component of an application system. 

We believe that decontextualization will also be helpful in a wide range of other applications. 
For example, in multi-document summarization~\cite{Fabbri2019MultiNewsAL}, co-referring entities and events must be resolved across different documents and removing ambiguous references may help; extractive summarization \cite{Cheng2016NeuralSB} could benefit from the type of pre-processing that we presented for open-domain QA; anaphora resolution is crucial for both summarization and machine translation~\cite{Susanne1992AnaphoraRI}; and decontextualizing sentences may help in recovering explicit mentions of entities and relations which can help information extraction~\cite{Narasimhan2016ImprovingIE}. 
The current formulation focuses on the English encyclopedic corpus and rewriting for an empty context, and future work can explore different domains of text as well as mapping to a different context.